\documentclass[11pt,a4paper]{article}
\usepackage[hyperref]{acl2020}

\usepackage{times}
\usepackage{latexsym}
\usepackage{algorithm}
\usepackage{algpseudocode}
\usepackage{color}
\usepackage{graphicx}
\usepackage{multirow}
\usepackage{paralist}
\usepackage{pifont}
\usepackage{subfigure}
\usepackage{url}
\usepackage{xspace}
\usepackage{amsfonts,amsmath,balance}
\usepackage{booktabs}
\usepackage{comment}

\usepackage{url}

\aclfinalcopy % Uncomment this line for the final submission
\newcommand*{\Scale}[2][4]{\scalebox{#1}{$#2$}}
\definecolor{darkgreen}{rgb}{0.0,0.4,0.0}
\definecolor{darkblue}{rgb}{0.0,0.0,0.6}
\definecolor{darkred}{rgb}{0.7,0.0,0.0}

\newcommand{\textdarkred}[1]{\textcolor{darkred}{#1}}
\newcommand{\textdarkblue}[1]{\textcolor{darkblue}{#1}}
\newcommand{\textdarkgreen}[1]{\textcolor{darkgreen}{#1}}

\newcommand{\highblue}[1]{\textdarkblue{\underline{#1}}}
\newcommand{\highorange}[1]{\textdarkred{\underline{#1}}}
\newcommand{\highgreen}[1]{\textdarkgreen{\underline{#1}}}

\title{Crossing Variational Autoencoders for Answer Retrieval}

\begin{document}

\author{Wenhao Yu$^{\dag}$, Lingfei Wu$^{\ddag}$, Qingkai Zeng$^{\dag}$, Shu Tao$^{\ddag}$, Yu Deng$^{\ddag}$,  Meng Jiang$^{\dag}$ \\
$\dag$University of Notre Dame, Notre Dame, IN, USA\\ 
$\ddag$IBM Thomas J. Watson Research Center, Yorktown Heights, NY, USA\\
{\tt $\dag$\{wyu1, qzeng, mjiang2\}@nd.edu} \\
{\tt $\ddag$\{wuli, shutao, dengy\}@us.ibm.com}
}

\maketitle
\begin{abstract}

Answer retrieval is to find the most aligned answer from a large set of candidates given a question. Learning vector representations of questions/answers is the key factor. Question-answer alignment and question/answer semantics are two important signals for learning the representations. Existing methods learned semantic representations with dual encoders or dual variational auto-encoders. The semantic information was learned from language models or question-to-question (answer-to-answer) generative processes. However, the alignment and semantics were too separate to capture the \textit{aligned semantics} between question and answer. In this work, we propose to \textit{cross} variational auto-encoders by generating questions with aligned answers and generating answers with aligned questions. Experiments show that our method outperforms the state-of-the-art answer retrieval method on SQuAD.
% improves MRR and R@1 by 1.7\% and 4.6\% over
% respectively.

% Retrieving relevant answers from a huge corpus of documents have been extensively researched, which typically finds the relevant articles first and identifies the answer spans from returned articles. In practice, sentence is a good size to present a user with a detailed answer, making it unnecessary to highlight specific spans for many use cases. However, the challenge of sentence-level retrieval lies in that traditional dual encoder models lack adequate representation capacity, making it difficult to determine answerability in specific perspective, especially when one answer corresponds to multiple questions from different perspectives. In this paper, we propose Crossing Variational Autoendcoders (CrossVAE) to strengthen the alignment of questions and answers through cross construction. Experiments demonstrate our proposed methods with lightweight language model (e.g., one-layer LSTM) could outperform state-of-art methods, which improves MRR and R@1 by 1.73\% and 4.59\% on SQuAD, respectively.

%  due to the short length of answers

\end{abstract}

\section{Introduction}
\label{sec:introduction}
% Given a question, the expected answer can be a word span~\cite{chen2017reading,danda2018too} or a sentence of detailed information~\cite{seo2019real,ahmad2019reqa}. Sentence-level answer retrieval approaches rely on learning vector representations (i.e., embeddings) of questions and answers from training data of pairs of question-answer texts. The question-answer alignment and question/answer semantics are expected to be preserved in the representations. In other words, the question/answer embeddings must reflect their semantics in the texts of being aligned as pairs.

Answer retrieval is to find the most aligned answer from a large set of candidates given a question~\cite{ahmad2019reqa,abbasiyantaeb2020text}. It has been paid increasing attention by the NLP and information retrieval community~\cite{yoon2019compare,chang2020pre}. Sentence-level answer retrieval approaches rely on learning vector representations (i.e., embeddings) of questions and answers from pairs of question-answer texts. The question-answer alignment and question/answer semantics are expected to be preserved in the representations. In other words, the question/answer embeddings must reflect their semantics in the texts of being aligned as pairs.

\begin{table}[]
\centering
\caption{The answer at the bottom of this table was aligned to 17 different questions at the sentence level.}
\label{tab:stoa-result}
\scalebox{0.88}{%
\linespread{1.1}
\begin{tabular}{p{8.2cm}}
\toprule
\textbf{Question (1):} What \highblue{three stadiums} did the NFL decide between for the game? \\
\textbf{Question (2):} What \highgreen{three cities} did the NFL consider for the game of Super Bowl 50? \\
\centerline{...} 
\textbf{Question (17):} \highorange{How many sites} did the NFL narrow down Super Bowl 50's location to? \\ \hline
\textbf{Answer:} The league eventually narrowed the bids to \highorange{three} sites: \highgreen{New Orleans} \highblue{Mercedes-Benz Superdome}, \highgreen{Miami} \highblue{Sun Life Stadium}, and the \highgreen{San Francisco} Bay Area's \highblue{Levi's Stadium}. \\ 
\bottomrule
\end{tabular}}
\label{intro}
\vspace{-0.2in}
\end{table}

\begin{figure*}[t]
	\centering
	\subfigure[Dual-Encoders~\cite{yang2019multilingual}]
	{\includegraphics[width=0.32\textwidth]{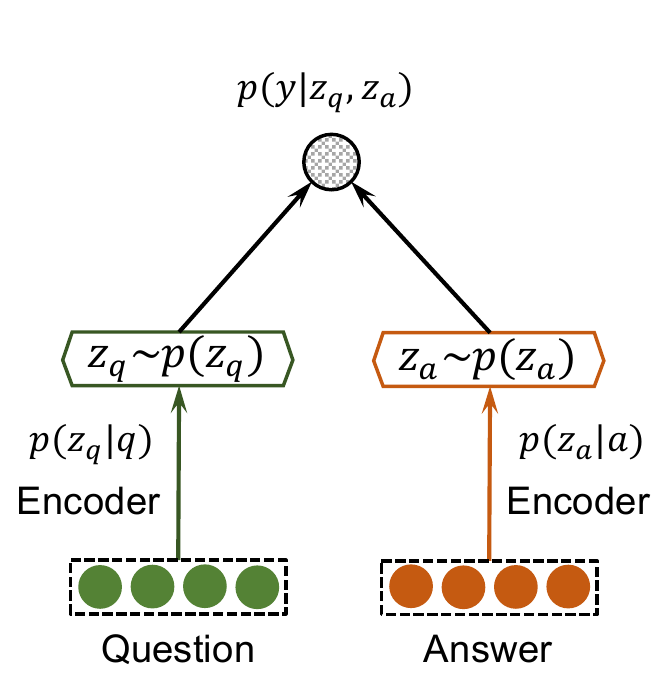}\label{frame-a}}
	\subfigure[Dual-VAEs~\cite{shen2018deconvolutional}]
	{\includegraphics[width=0.32\textwidth]{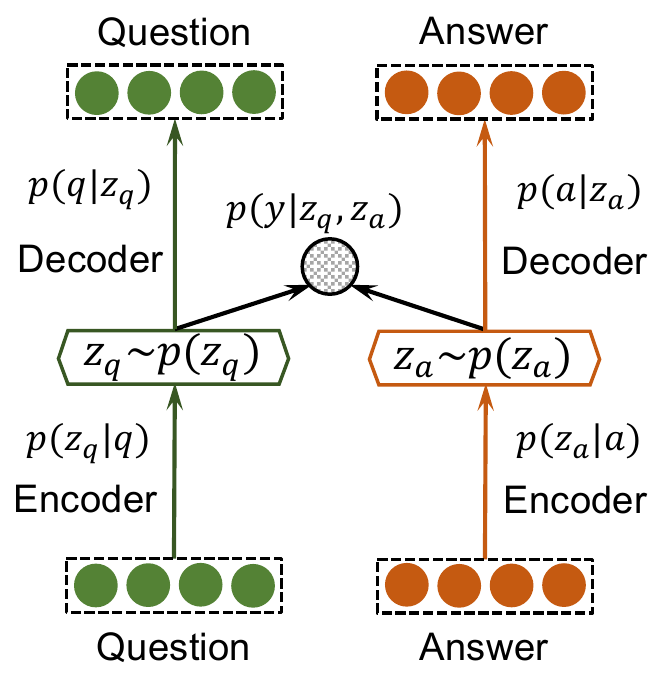}\label{frame-b}}
	\subfigure[Dual-CrossVAEs (\textbf{Ours})]
	{\includegraphics[width=0.32\textwidth]{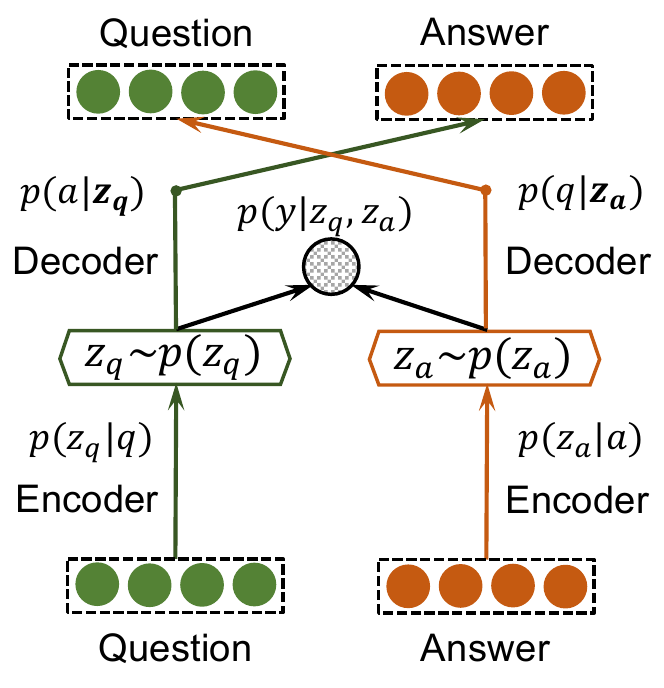}\label{frame-c}}
	\vspace{-0.1in}
	\caption{(a)--(b) The Q-A alignment and Q/A semantics were learned too separately to capture the aligned semantics between question and answer. (c) We propose to cross VAEs by generating questions with aligned answers and generating answers with aligned questions. }
	\label{framework}
	\vspace{-0.15in}
\end{figure*}

One popular scheme ``Dual-Encoders'' (also known as ``Siamese network'' ~\cite{triantafillou2017few,das2016together}) has two separate encoders to generate question and answer embeddings and a predictor to match two embedding vectors~\cite{cer2018universal,yang2019multilingual}. Unfortunately, it has been shown difficult to train deep encoders with the weak signal of matching prediction~\cite{bowman2015large}. Then there has been growing interests in developing deep generative models such as variational auto-encoders (VAEs) and generative adversial networks (GANs) for learning text embeddings~\cite{xu2017variational,xie2019dual}. As shown in Figure \ref{frame-b}, the scheme of ``Dual-VAEs'' has two VAEs, one for question and the other for answer~\cite{shen2018deconvolutional}. It used the tasks of generating reasonable question and answer texts from latent spaces for preserving semantics into the latent representations.

Although Dual-VAEs was trained jointly on question-to-question and answer-to-answer reconstruction, the question and answer embeddings can only preserve \textit{isolated} semantics of themselves. In the model, the Q-A alignment and Q/A semantics were too separate to capture the \textit{aligned semantics} (as we mentioned at the end of the first paragraph) between question and answer. Learning the alignment with the weak Q-A matching signal, though now based on generatable embeddings, can lead to confusing results, when (1) different questions have similar answers and (2) similar questions have different answers. Table~\ref{intro} shows an examples in SQuAD: 17 different questions share the same sentence-level answer.

Our idea is that if aligned semantics were preserved, the embeddings of a question would be able to generate its answer, and the embeddings of an answer would be able to generate the corresponding question. In this work, we propose to \textit{cross} variational auto-encoders, shown in Figure \ref{frame-c}, by reconstructing answers from question embeddings and reconstructing questions from answer embeddings. Note that compared with Dual-VAEs, the encoders do not change but decoders work across the question and answer semantics.

Experiments show that our method improves MRR and R@1 over the state-of-the-art method by 1.06\% and 2.44\% on SQuAD, respectively. On a subset of the data where any answer has at least 10 different aligned questions, our method improves MRR and R@1 by 1.46\% and 3.65\%, respectively.

\section{Related Work}
\label{sec:RelatedWork}
Answer retrieval (AR) is defined as the answer of a candidate question is obtained by finding the most similar answer between multiple candidate answers~\cite{abbasiyantaeb2020text}. While another popular task on SQuAD dataset is machine reading comprehension (MRC), which is introduced to ask the machine to answer questions based on one given context~\cite{liu2019neural}. In this section, we review existing work related to answer retrieval and variational autoencoders.

\vspace{0.05in}
\noindent \textbf{Answer Retrieval.} It has been widely studied with information retrieval techniques and has received increasing attention in the recent years by considering deep neural network approaches. Recent works have proposed different deep neural models in text-based QA which compares two segments of texts and produces a similarity score. Document-level retrieval ~\cite{chen2017reading,wu2018word,seo2018phrase,seo2019real} has been studied on many public datasets including including SQuAD~\cite{rajpurkar2016squad}, MsMarco~\cite{nguyen2016ms} and NQ~\cite{kwiatkowski2019natural} etc. ReQA proposed to investigate sentence-level retrieval and provided strong baselines over a reproducible construction of a retrieval evaluation set from the SQuAD data~\cite{ahmad2019reqa}. We also focus on sentence-level answer retrieval.

\vspace{0.05in}
\noindent \textbf{Variational Autoencoders.} VAE consists of encoder and generator networks which encode a data example to a latent representation and generate samples from the latent space, respectively~\cite{kingma2013auto}. Recent advances in neural variational inference have manifested deep latent-variable models for natural language processing tasks~\cite{bowman2016generating,kingma2016improved,hu2017toward,hu2017unifying,miao2016neural}. The general idea is to map the sentence into a continuous latent variable, or code, via an inference network (encoder), and then use the generative network (decoder) to reconstruct the input sentence conditioned on samples from the latent code (via its posterior distribution). Recent work in cross-modal generation adopted cross alignment VAEs to jointly learn representative features from multiple modalities~\cite{liu2017unsupervised,shen2017style,schonfeld2019generalized}. DeConv-LVM~\cite{shen2018deconvolutional} and VAR-Siamese~\cite{deudon2018learning} are most relevant to us, both of which adopt Dual-VAEs models (see Figure \ref{frame-b}) for two text sequence matching task. In our work, we propose a Cross-VAEs for questions and answers alignment to enhance QA matching performance.

\section{Proposed Method}
\noindent\textbf{Problem Definition.} Suppose we have a question set $\mathcal{Q}$ and an answer set $\mathcal{A}$. Each question and answer have only one sentence. Each question $q \in \mathcal{Q}$ and answer $a \in \mathcal{A}$ can be represented as $(q, a, y)$, where $y$ is a binary variable indicating whether $q$ and $a$ are aligned. Therefore, the solution of sentence-level retrieval task could be considered as a matching problem.
Given a question $q$ and a list of answer candidates $C(q) \subset \mathcal{A}$, 
% In the input data, each question $q \in Q$ comes with a list of an associated answer candidates set $C(q) \subset \mathcal{A}$.
our goal is to predict $p(y|q, a)$ of each input question $q$ with each answer candidate $a \in C(q)$.
% and to generate an optimal ranking $R$ sorting all probabilities $p(y|q, a)$ from highest to lowest.

\subsection{Crossing Variational Autoencder}

Learning cross-domain constructions under generative assumption is essentially learning the conditional distribution $p(q|z_a)$ and $p(a|z_q)$ where two continuous latent variables $z_{q}, z_{a} \in \mathbb{R}^{d_z}$ are independently sampled from $p(z_q)$ and $p(z_a)$:
\begin{equation}
    p(q|a) = \mathbb{E}_{z_{a}\sim p(z_{a}|a)}[p(q|z_{a})],
\end{equation}
\vspace{-0.2in}
\begin{equation}
    p(a|q) = \mathbb{E}_{z_{q}\sim p(z_{q}|q)}[p(a|z_{q})].
\end{equation}
The question-answer pair matching can be represented as the conditional distribution $p(y|z_q, z_a)$ from latent variables $p(q|z_a)$ and $p(a|z_q)$:
\begin{equation}
    \Scale[0.94] {p(y|q,a) = \mathbb{E}_{z_{q}\sim p(z_{q}|q), z_{a}\sim p(z_{a}|a)}}[p(y|z_q, z_a)],
\end{equation}

\vspace{0.05in}
\noindent\textbf{Objectives.} We denote $E_q$ and $E_a$ as question and answer encoders that infer the latent variable $z_q$ and $z_a$ from a given question answer pair $(q, a, y)$, and $D_q$ and $D_a$ as two different decoders that generate corresponding question and answer $q$ and $a$ from latent variables $z_a$ and $z_q$. Then, we have cross construction objective function:

\vspace{-0.05in}
\begin{equation}
\begin{aligned}
    \mathcal{L}_{cross} (\theta_{E}, & \theta_{D}) \\ = & y \cdot \mathbb{E}_{q\sim Q}[-\log p_{D}(q|a, E(a))] \\ + & {y \cdot \mathbb{E}_{a\sim A}[-\log p_{D}(a|q, E(q))].}
\end{aligned}
\end{equation}
Variational Autoencoder~\cite{kingma2013auto} imposes KL-divergence regularizer to align both posteriors $p_E(z_q|q)$ and $p_E(z_a|a)$:

\begin{equation}
\begin{split}
    \Scale[0.95] {\mathcal{L}_{KL} (\theta_{E}) = } & \Scale[0.95] {y \cdot \mathbb{E}_{q \sim Q}[{D_{KL}}(p_{E}(z_q|q)||p(z_q))] }\\ + & \Scale[0.95] {y \cdot \mathbb{E}_{a \sim A}[{D_{KL}}(p_{E}(z_a|a)||p(z_a))],}
\end{split}
\end{equation}
where $\theta_{E}$, $\theta_{D}$ are all parameters to be optimized. Besides, we have question answer matching loss from $f_\phi(y|q,a)$ as:

\begin{equation}
\begin{aligned}
     \mathcal{L}_{matching} & (\phi_{f}) =  -\big{[}y \cdot \log p_{f_\phi}(y|z_q, z_a) \\ + & (1-y) \cdot \log(1-p_{f_\phi}(y|z_q, z_a))\big{]},
\end{aligned}
\end{equation}
where $f$ is a matching function and $\phi_f$ are parameters to be optimized. Finally, in order to allow the model to balance between maximizing the variational evidence lower bound (ELBO) and minimizing the question answer matching loss, a joint training objective is given by:

\vspace{-0.05in}
\begin{equation}
    \mathcal{J} = - \alpha \cdot \mathcal{L}_{cross} - \beta \cdot \mathcal{L}_{KL} + \gamma \cdot \mathcal{L}_{matching},
\end{equation}
where $\alpha$, $\beta$ and $\gamma$ are introduced as  hyper-parameters to control the importance of each task.

\subsection{Model Implementation}

\vspace{0.05in}
\noindent\textbf{Dual Encoders.} We use Gated Recurrent Unit (GRU) as encoders to learn contextual words embeddings~\cite{cho2014learning}. Question and answer embeddings are reduced by weighted sum through multiple hops self-attention~\cite{lin2017structured} of GRU units and then fed into two linear transition to obtain mean and standard deviation as $\mathcal{N}(z_q;\mu_q,diag(\sigma_q^2))$ and $\mathcal{N}(z_a;\mu_a,diag(\sigma_a^2))$.

\vspace{0.05in}
\noindent\textbf{Dual Decoders.} We adopt another Gated Recurrent Unit (GRU) for generating token sequence conditioned on the latent variables $z_q$ and $z_a$. 

\vspace{0.05in}
\noindent\textbf{Question Answer Matching.}  We adopt cosine similarity with $l_2$ normalization to measure the matching probability of a question answer pair.

\section{Experiment}
\subsection{Dataset}
Our experiments were conducted on SQuAD 1.1~\cite{rajpurkar2016squad}.
% since SQuAD 2.0 adds questions which might not have a corresponding answer in the paragraph. 
It has over 100,000 questions composed to be answerable by text from Wikipedia documents. Each question has one corresponding answer sentence extracted from the Wikipedia document. Since the test set is not publicly available, we partition the dataset into 79,554 (training) / 7,801 (dev) / 10,539 (test) objects. 

\begin{table}[t]
\centering
\caption{Performance of answer retrieval on SQuAD.}
\label{tab:stoa-result}
\setlength{\tabcolsep}{4.5mm}{
\scalebox{0.88}{%
\begin{tabular}{l|ccc}
\toprule
\multirow{2}{*}{\textbf{Method}} & \multicolumn{3}{c}{\textbf{SQuAD}} \\
 & MRR  & R@1 & R@5 \\ 
\midrule
\textbf{InferSent} & 36.90  & 27.91 & 46.92  \\
\textbf{SenBERT}  & 38.01 & 27.34 & 49.59 \\
\textbf{BERT}$\bf_{QA}$ & 48.07  & 40.63 & 57.45 \\
\textbf{QA-Lite} & 50.29 & 40.69 & 61.38 \\
\textbf{USE-QA}  & 61.23 & 53.16 & 69.93 \\ 
\midrule
\textbf{Dual-GRUs}  & 61.06  & 54.70 & 68.25  \\
\textbf{Dual-VAEs}  & 61.48 & 55.01 & 68.49 \\
\textbf{Cross-VAEs}  & \textbf{62.29}  & \textbf{55.60} &\textbf{70.05 } \\
\bottomrule
\end{tabular}}}
\label{baselines}
\end{table}

\begin{table}[t]
\centering
\caption{Performance of answer retrieval on a subset of SQuAD in which any answer has more than 8 questions. Our method outperforms baselines much more. SSE indicates the sum of squared distances/errors between two different questions aligned to same answer.}
\label{tab:stoa-result}
\setlength{\tabcolsep}{3.2mm}{
\scalebox{0.88}{%
\begin{tabular}{l|cccc}
\toprule
\multirow{2}{*}{\textbf{Method}} & \multicolumn{4}{c}{\textbf{SQuAD Subset}} \\
 & MRR & R@1 & R@5 & SSE \\ 
 \midrule
\textbf{BERT}$\bf_{QA}$ & 37.90 & 30.81 & 45.24 & 0.23 \\
\textbf{USE-QA} & 47.06 & 40.90 & 53.44 &  0.14 \\
\textbf{Cross-VAEs} & \textbf{48.52} & \textbf{44.55} & \textbf{53.52} & \textbf{0.09}\\
\bottomrule
\end{tabular}}}
\label{subset}
\end{table}

\subsection{Baselines}
% Dual encoder architecture has shown strong performance on tasks of conversational response retrieval~\cite{yang2018learning,henderson2017efficient}, translation pair retrieval~\cite{guo2018effective,yang2019improving} and similar text retrieval~\cite{gillick2018end}.

\vspace{0.05in}
\noindent \textbf{InferSent}~\cite{conneau2017supervised}\textbf{.} It is not explicitly designed for answer retrieval,
% \footnote{The non-QA versions of the Universal Sentence Encoder produce general semantic embeddings of text.},
but it produces results on semantic tasks without requiring additional fine tuning.

\vspace{0.05in}
\noindent \textbf{USE-QA}~\cite{yang2019multilingual} \textbf{.}
% \footnote{Codes are published by Google, which could be downloaded at \url{https://tfhub.dev/google/universalsentence-encoder-multilingual-qa}}.}
It is based on Universal Sentence Encoder~\cite{cer2018universal}, but trained with multilingual QA retrieval and two other tasks: translation ranking and natural language inference. The training corpus contains over a billion question answer pairs from popular online forums and QA websites (e.g, Reddit).

\vspace{0.05in}
\noindent \textbf{QA-Lite.} Like USE-QA, this model is also trained over online forum data based on transformer. The main differences are reduction in width and depth of model layers, and sub-word vocabulary size.

\vspace{0.05in}
\noindent \textbf{\textbf{BERT}$\bf_{QA}$}~\cite{devlin2019bert} \textbf{.}
BERT$_{QA}$ first concatenates the question and answer into a text sequence $[[CLS], Q, [SEP], A, [SEP]]$, then passes through a 12-layers BERT and takes the $[CLS]$ vector as input to a binary classifier.

\vspace{0.05in}
\noindent \textbf{\textbf{SenBERT}}~\cite{reimers2019sentence} \textbf{.}
It consists of twin structured BERT-like encoders to represent question and answer sentence, and then applies a similarity measure at the top layer.

\subsection{Experimental Settings}

\vspace{0.05in}
\noindent \textbf{Implementation details.} We initialize each word with a 768-dim BERT token embedding vector. If a word is not in the vocabulary, we use the average vector of its sub-word embedding vectors in the vocabulary. The number of hidden units in GRU encoder are all set as 768. All decoders are multi-layer perceptions (MLP) with one 768 units hidden layer. The latent embedding size is 512. The model is trained for 100 epochs by SGD using Adam optimizer~\cite{kingma2014adam}. For the KL-divergence, we use an KL cost annealing scheme~\cite{bowman2016generating}, which serves the purpose of letting the VAE learn useful representations before they are smoothed out. We increase the weight $\beta$ of the KL-divergence by a rate of $2/epochs$ per epoch until it reaches 1. We set learning rate as 1e-5, and implemented on Pytorch.

\vspace{0.05in}
\noindent \textbf{Competitive Methods.} We compare our proposed method cross variational autoencoder (Cross-VAEs) with dual-encoder model and dual variational autoencoder (Dual-VAEs). For fair comparisons, we all use GRU as encoder and decoder, and keep all other hyperparameters the same.

\vspace{0.05in}
\noindent \textbf{Evaluation Metrics.}
The models are evaluated on retrieving and ranking answers to questions using three metrics, mean reciprocal rank (MRR) and recall at K (R@K). R@K is the percentage of correct answers in topK out of all the relevant answers. MRR represents the average of the reciprocal ranks of results for a set of queries.

\begin{figure*}[t]
    \centering
    \subfigure[USE-QA]
    {\includegraphics[width=0.46\textwidth]{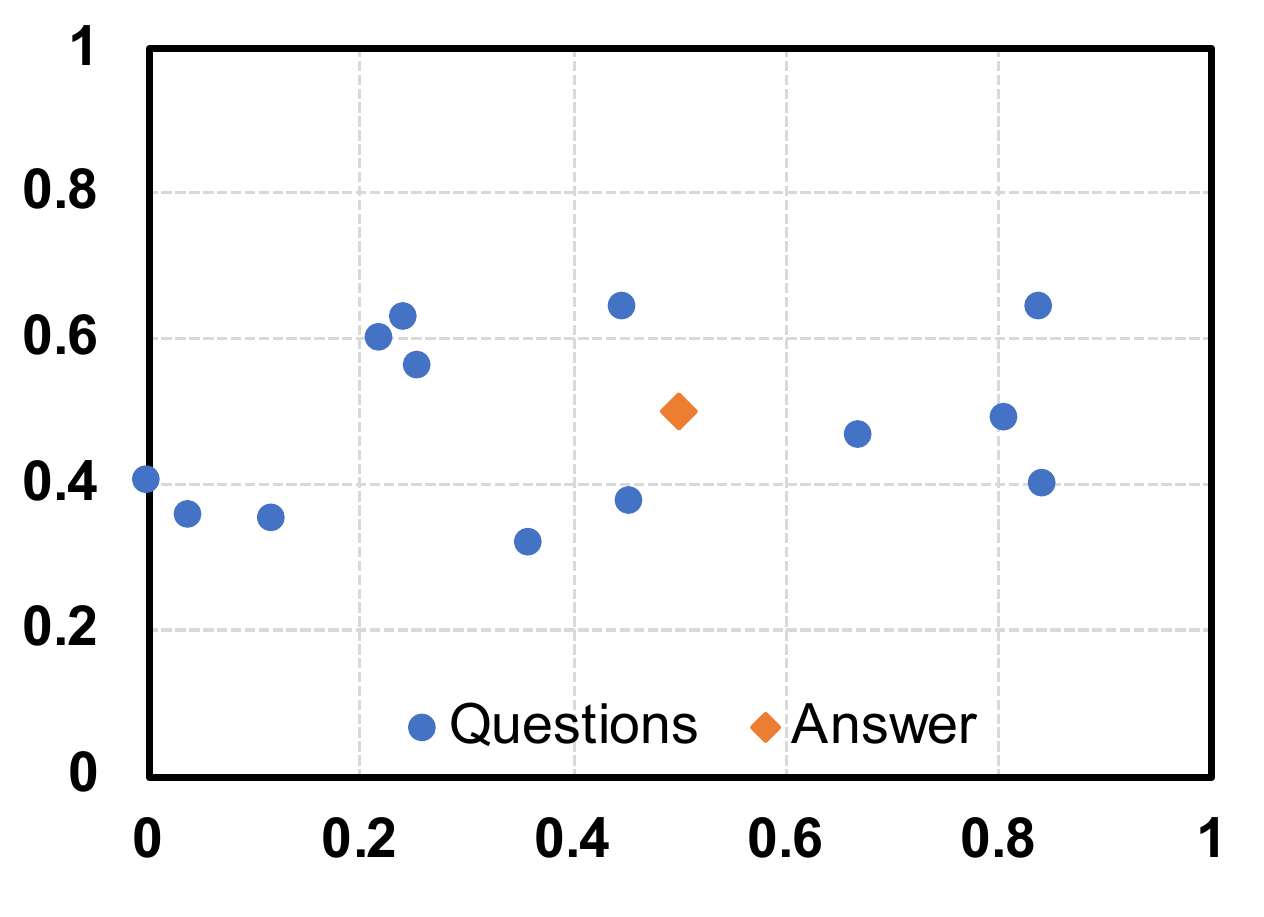}\label{fig:template11}}
    \subfigure[CrossVAEs]
    {\includegraphics[width=0.46\textwidth]{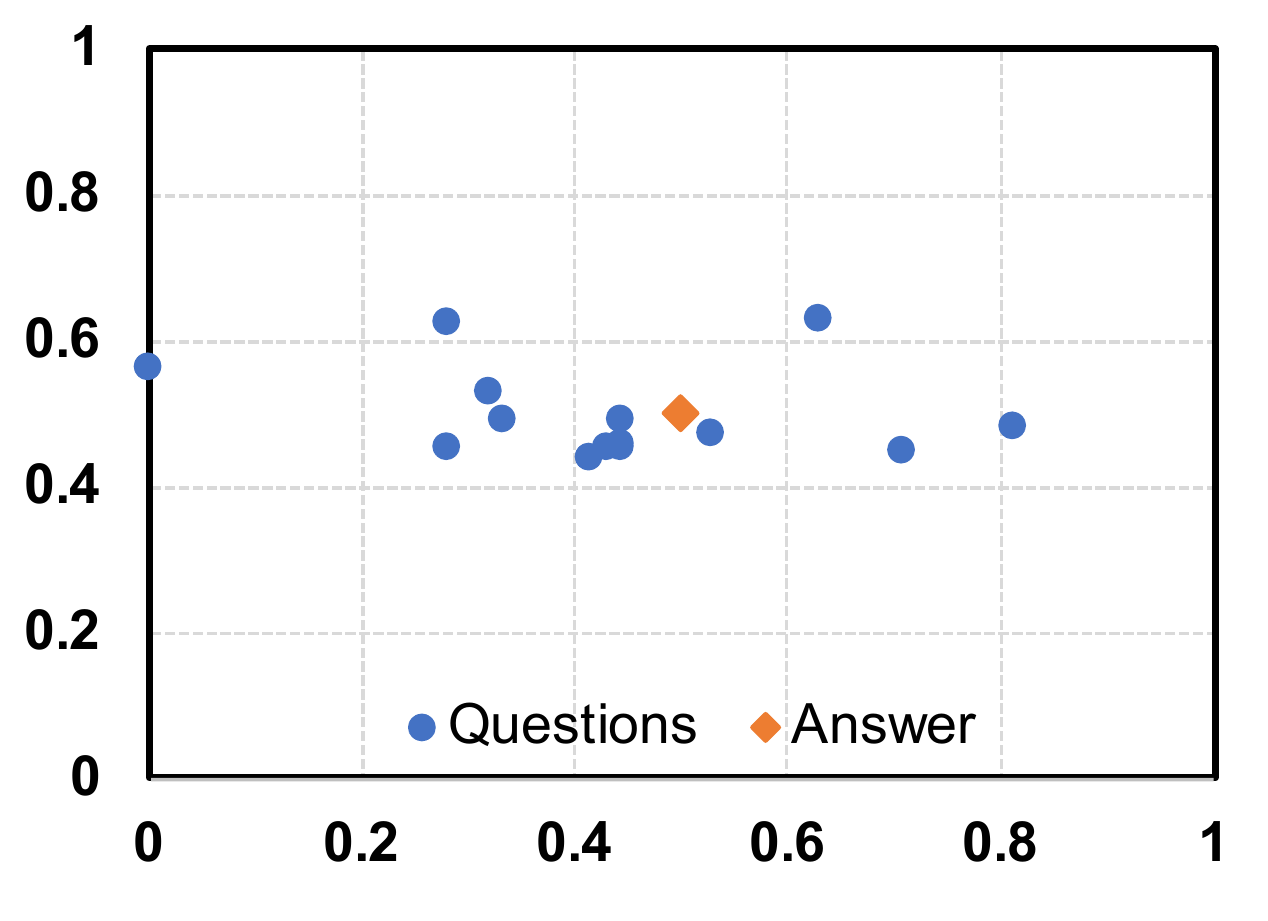}\label{fig:template12}}
    \subfigure[Two questions were incorrectly matched by USE-QA, but correctly matched by CrossVAEs.]
    {\includegraphics[width=0.94\textwidth]{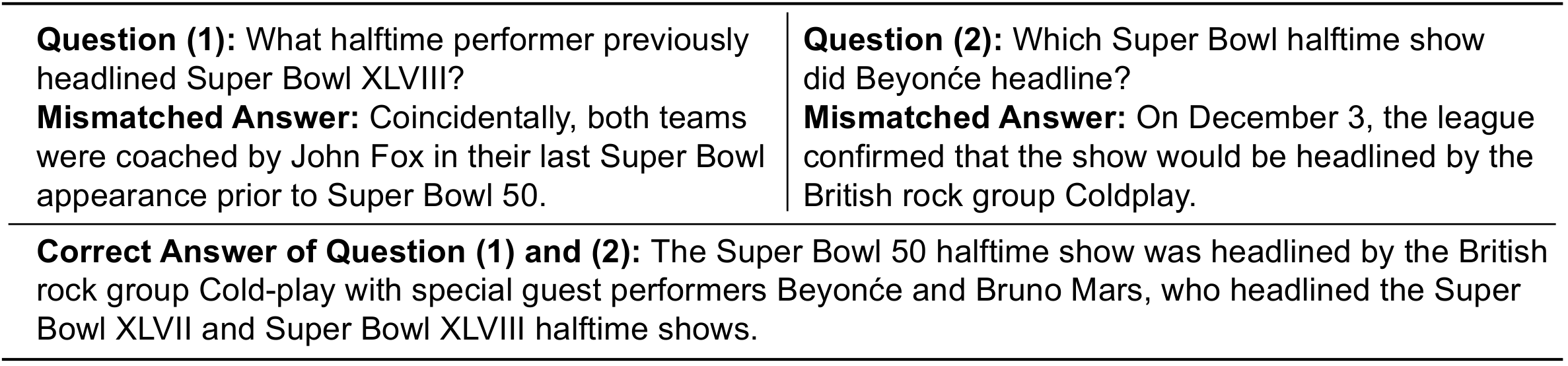}\label{fig:template13}}
    \vspace{-0.1in}
    \caption{A case of 14 different questions aligned to the same answer. We use SVD to reduce embedding dimensions to 2, and then project them on the X-Y coordinate axis. The scale of X-Y axis is relative with no practical significance. We observe that our method makes questions that share the same answer to be closer with each other.}
    \label{case}
\end{figure*}

\vspace{0.05in}
\noindent\textbf{Comparing performance with baselines.}
As shown in Table \ref{baselines}, two BERT based models do not perform well, which indicates fune tuning BERT may not be a good choice for answer retrieval task due to unrelated pre-training tasks (e.g, masked language model). In contrast, using BERT token embedding can perform better in our retrieval task. Our proposed method outperforms all baseline methods. Comparing with USE-QA, our method improves MRR and R@1 by +1.06\% and +2.44\% on SQuAD, respectively. In addition, Dual variational autoencoder (Dual-VAEs) does not make much improvement on question answering retrieval task because it can only preserve isolated semantics of themselves. Our proposed crossing variational autoencoder (Cross-VAEs) could outperform dual-encoder model and dual variational autoencoder model, which improves MRR and R@1 by +1.23\%/+0.81\% and +0.90\%/+0.59\%, respectively.

\vspace{0.05in}
\noindent\textbf{Analyzing performance on sub-dataset.} We extract a subset of SQuAD, in which any answer has at least eight different questions. As shown in Table \ref{subset}, our proposed cross variational autoencoder (Cross-VAEs) could outperform baseline methods on the subset. Our method improves MRR and R@1 by +1.46\% and +3.65\% over USE-QA. Cross-VAEs significantly improve the performance when an answer has multiple aligned questions. Additionally, SSE of our method is smaller than that of USE-QA. Therefore, the questions of the same answer are closer in the latent space.

\subsection{Case Study}

Figures \ref{fig:template11} and \ref{fig:template12} visualize embeddings of 14 questions of the same answer. We observe that crossing variational autoencoders (CrossVAE) can better capture the aligned semantics between questions and answers, making latent representations of questions and answers more prominent. Figure \ref{fig:template13} demonstrates two of example questions and corresponding answers produced by USE-QA and CrossVAEs. We observe that CrossVAEs can better distinguish similar answers even though they all share several same words with the question.

\section{Conclusion}
Given a candidate question, answer retrieval aims to find the most similar answer text between candidate answer texts. In this paper, We proposed to cross variational autoencoders by generating questions with aligned answers and generating answers with aligned questions. Experiments show that our method improves MRR and R@1 over the best baseline by 1.06\% and 2.44\% on SQuAD.

\section*{Acknowledgements}

% The authors would like to thank the anonymous referees for their valuable comments and helpful suggestions. 
We thank Drs. Nicholas Fuller, Sinem Guven, and Ruchi Mahindru for their constructive comments and suggestions. This project was partially supported by National Science Foundation (NSF) IIS-1849816 and Notre Dame Global Gateway Faculty Research Award.

\balance
\bibliography{references}
\bibliographystyle{acl_natbib}

\end{document}